\documentclass[journal]{IEEEtran}
\usepackage{mathrsfs,amsmath}
\usepackage{stmaryrd}
\usepackage{amsfonts}
\usepackage{algorithm,algorithmic}
\usepackage{subfigure}
\usepackage{verbatim}
\usepackage{cite}
\usepackage{blindtext}
\usepackage{amssymb}
\usepackage{esint,caption}
\usepackage{graphicx,xcolor,subfig}
\usepackage[section] {placeins}
\usepackage{morefloats}
\graphicspath{{./pics/}}
\usepackage{eepic,epic,epsfig,epstopdf}
\usepackage{url}

\ifCLASSINFOpdf
\else
\fi

\begin{document}

\title{A stochastic alternating minimizing method for sparse phase retrieval }
\author{Jian-Feng Cai, Yuling Jiao, Xiliang Lu,
        and Juntao You
\thanks{Copyright (c) 2017 IEEE. Personal use of this material is permitted. However, permission to use this material for any other purposes must be obtained from the IEEE by sending a request to pubs-permissions@ieee.org \protect\\
Jianfeng Cai is in the  Department of Mathematics, The Hong Kong University of Science and Technology Kowloon, Hong Kong, (email: jfcai@ust.hk), Yuling Jiao is in the School of Statistics and Mathematics, Zhongnan University of Economics and Law, Wuhan, 430063, P.R. China (email: yulingjiaomath@whu.edu.cn), Xiliang Lu (corresponding author) is in the School of Mathematics and Statistics, Wuhan University and
Hubei Key Laboratory of Computational Science, Wuhan University, Wuhan 430072, P.R. China (email: xllv.math@whu.edu.cn) and
Juntao You is in  the  Department of Mathematics, The Hong Kong University of Science and Technology Kowloon, Hong Kong, (email: jyouab@connect.ust.hk).
}}

\markboth{IEEE SIGNAL PROCESSING Letters,Vol.~~, No.~~, ~~,2019}%
{Shell \MakeLowercase{\textit{et al.}}: Bare Demo of IEEEtran.cls for Journals}
\maketitle

\begin{abstract}
Spares phase retrieval plays an important role in many fields of applied science and thus attracts lots of attention.
In this paper, we propose a \underline{sto}chastic alte\underline{r}nating \underline{m}inimizing method for \underline{sp}arse ph\underline{a}se \underline{r}etrieval      (\textit{StormSpar}) algorithm which {emprically} is able to recover  $n$-dimensional $s$-sparse signals from only $O(s\,\mathrm{log}\, n)$ number of measurements without a desired initial value required by many existing methods.
 In  \textit{StormSpar},  the hard-thresholding pursuit (HTP) algorithm is employed to solve the sparse constraint least square sub-problems. The main competitive feature of \textit{StormSpar} is   that it converges globally   requiring optimal order of  number of samples with random initialization.  Extensive numerical experiments are given to validate the proposed algorithm.
\end{abstract}


\markboth{IEEE TRANSACTIONS ON INDUSTRIAL ELECTRONICS}%
{}

\definecolor{limegreen}{rgb}{0.2, 0.8, 0.2}
\definecolor{forestgreen}{rgb}{0.13, 0.55, 0.13}
\definecolor{greenhtml}{rgb}{0.0, 0.5, 0.0}

\section{Introduction}

\IEEEPARstart{P}{hase} retrieval is to recover the phase information from its magnitude measurements, i.e.,
\begin{align} \label{problem}
y_i=| \langle a_i,x \rangle |+ \epsilon_i , \quad i=1,2,\cdots,m,
\end{align}
where $x\in \mathbb{F} ^n$ is the unknown vector, $a_i \in \mathbb{F} ^n$ are given sampling vectors which are  random Gaussian vector in this paper, $y_i$ is the observed measurements, $\epsilon_i$ is the noise, and $m$ is the number of measurements (or the sample size). The $\mathbb{F} ^n$ can be $\mathbb{R} ^n$ or $\mathbb{C} ^n$, and we consider the real case $\mathbb{F} ^n=\mathbb{R} ^n$ in this work.
The phase retrieval problem arises in many fields like X-ray crystallography \cite{harrison1993phase}, optics \cite{walther1963question}, microscopy \cite{miao2008extending} and others, see e.g., \cite{fienup1982phase}. Due to the lack of phase information, the phase retrieval problem is a nonlinear and ill-posed problem.

When the measurements are overcomplete i.e., $m>n$, there are many algorithms in the literatures. Earlier approaches were mostly based on alternating projections, e.g. the work of Gerchberg and Saxton \cite{gerchberg1972practical} and Fienup \cite{fienup1982phase}. Recently, convex relaxation methods such as phase-lift \cite{candes2015phase} and phase-cut \cite{waldspurger2015phase} have been proposed. These methods transfer the phase retrieval problem into a semi-definite programing, 
which can be computationally expensive. Another convex approach named phase-max which does not lift the dimension of the signal was proposed in  \cite{goldstein2016phasemax}. In the mean while, there are other works based on solving  nonconvex optimization via first and second order methods including  alternating minimization \cite{netrapalli2013phase} (or Fienup methods), Wirtinger flow \cite{candes2015phase}  Kaczmarz \cite{wei2015solving}, Riemannian optimization \cite{cai2018solving};
Gauss-Newton \cite{gao2016gauss,gaussnewton18} etc. With a good initialization abstained via spectral methods, the above mentioned methods work with theoretical guarantees. Progresses have been made by replacing the desired initialization  with random initialized ones in  alternating minimization \cite{adm18,zhang19}, gradient descent \cite{gradient18} and Kaczmarz method \cite{kaczmarz1,kaczmarz2}  while keeping  convergence guarantee
with high probability. {Also, recent analysis in \cite{sun2018geometrical,li2018towards} has shown that some nonconvex objective functions for phase retrieval have a nice landscape --- there is no spurious local minima --- with high probability. As a consequence, for these objective functions, any algorithms finding a local minima are guaranteed to give a successful phase retrieval.}

For the large scale problem, the requirement $m>n$ becomes unpractical due to the huge measurement and computation cost. In many applications, the true signal $x$ is known to be sparse. Then the sparse phase retrieval problem can be solved with a small number of sampling, thus possible to be applied to large scale problems.
It has been proved in \cite{eldar2014phase} that $m=O(s\mathrm{log}\, n/s)$ measurement is sufficient to ensure successful recovery in theory with high probability when the model is  Gaussian (i.e. the sampling vector $a_i$ are i.i.d  Gaussian and the  target   is real). But the exiting  computational trackable  algorithms require $O(s^{2}\mathrm{log}\, n)$ number of measurements to reconstruct the sparse signal, for example, $\ell_{1}$ regularized PhaseLift method \cite{li2013sparse}, sparse AltMin \cite{netrapalli2013phase}, GESPAR \cite{shechtman2014gespar}, Thresholding/projected Wirtinger flow \cite{cai2016optimal,wflowproj19},  SPARTA \cite{wang2016sparse} and so on. Two stage methods  based on phase-lift and compressing has been introduced in \cite{iwen2017robust,sparsephasenips}, which is able to do successful reconstruction with $O(s\mathrm{log}\, n)$ measurements for some special designed  sampling matrix which exclude the Gaussian model \eqref{problem}. When a good initialization is available, the sample complexity can be improved to $O(s\mathrm{log}\, n)$  \cite{phaselinear,wflowproj19}. However, it requires   $O(s^2\mathrm{log}\, n)$ samples to get a   desired  sparse  initialization in the existing literatures. This gap  naturally raises the following   challenging question

  \textit{Can one  recover the $s$-sparse target from the  phaseless generic Gaussian model \eqref{problem}  with  $O(s\mathrm{log}\, n)$ measurements  via just using  random initializations ?}

In this paper, we propose a novel algorithm to solve the sparse phase retrieval problems in the very limited measurements (numerical examples show that $m=O(s \mathrm{log} \, n)$ can be enough). The algorithm is a stochastic version of alternating minimizing method. The idea of alternating minimization method is: during each iteration, we first given an estimation of the phase information, then substitute the approximated phase into \eqref{problem} with the sparse constraint and solve a standard compressed sensing problem to get an updated sparse signal. But since the alternating minimizing method is a local method, it is very sensitive to the initialization. Without enough measurements, it is very difficult to compute a good initial guess. To overcome this difficult, we change the sample matrix during each iteration via bootstrap technique, see Algorithm \ref{alg:ithc} for details.
The numerical experiments shows that the proposed algorithm needs only $O(s \,\mathrm{log}\, n)$ measurements to recover the true signal with high probability in Gaussian model, and it works for a random initial guess. The experiments also show that the proposed algorithm is able to recover signal in a wide range of sparsity.

The rest of papers are as follows. In section II we will introduce the setting of problem and the details of the algorithm. Numerical experiments are given in section III.



\section{Algorithm}
First we introduce some notations. For any $a, b\in \mathbb{R}^{n}$, we denote that $a\odot b=(a_1 b_1,a_2 b_2,\cdots,a_n b_n)$, $\lVert x\lVert_0$ is the number of nonzero entries of $x$, and $\lVert x\lVert_2$ is the standard $\l _2$-norm, i.e. $\lVert x\lVert_2=\sqrt{\sum_{i=i}^{n}x_{i}^2}$. The floor function $\lfloor c \rfloor$ is the greatest integer which is less than or equal to $c$.

Recall from \eqref{problem}, We denote the sampling matrix and the measurement vector by {$A = [a_1^t; ...; a_m^t] \in \mathbb{R} ^{m\times n}$ and $y = [y_1; ...; y_m] \in \mathbb{R} ^m$}, respectively. Let $x\in \mathbb{R} ^n$ be the unknown sparse signal to be recovered. In the noise free case, the problem can be written as to find $x$ such that
 $$y=\lvert Ax \lvert, \quad \mathrm{s.t} \ \lVert x\lVert_0 \le s.$$
In the noise case, this can be written by the nonconvex minimization problem:
\begin{align}\label{least_sq}
\min_{x} \quad \frac{1}{2}\lVert y-| Ax |\lVert_{2}^{2}  \quad \mathrm{s.t.} \ \lVert x\lVert_{0} \le s \;,
\end{align}
Now we propose  the \underline{sto}chastic alte\underline{r}nating \underline{m}inimizing method for \underline{sp}arse ph\underline{a}se \underline{r}etrieval      (\textit{StormSpar}) as follows. It starts with a random initial guess $x^{0}$. In the $\ell$-th step of iteration ($\ell=1,2,\cdots$), we first randomly choose some rows of the sampling matrix $A$ to form a new matrix $A^{\ell}$ (which is a submatrix of $A$), and denoted by the corresponding rows of $y$ to $A^{\ell}$ is $y^{\ell}$.
Then we compute the phase information of  $A^{\ell}x^{\ell -1}$, say $p^{\ell} = \mathrm{sign}(A^{\ell}x^{\ell -1})$, and to solve the standard compressed sensing subproblem
\begin{equation}\label{pro:cs}
\min_{x} \frac{1}{2}\|A^{\ell} x - \tilde{y}^{\ell}\|^2  \quad \mathrm{s.t.} \ \lVert x\lVert_{0} \le s \; ,
\end{equation}
where $\tilde{y}^{\ell}=p^{\ell}\odot y^{\ell}$.  Problem \eqref{pro:cs} can be solved by a lot of compressed sensing solver, and we will use the efficient Hard Thresholding Pursuit (HTP) \cite{foucart2011hard} in our algorithm. For completion, HTP is given in Algorithm \ref{alg:htp}.
We summarize the StormSpar algorithm in the Algorithm \ref{alg:ithc}.

\section{Numerical Results and Discussions}

\subsection{Implementation details}
The true signal $x$ is chosen as $s$-sparse with random support and the design matrix $A\in\mathbb{R}^{m\times n}$ is chosen to be random Gaussian matrix. The additive gaussian noise following the form $\epsilon=\sigma*\textrm{randn}(n,1)$, thus the noise level is determined by $\sigma$. The parameter $\gamma$ is set to be $\mathrm{min}(\frac{s}{m}*\mathrm{log}\, \frac{n}{0.001},0.6)$, and $\delta = 0.01$.

The estimation error $r$ between the estimator $\hat{x}$ and the true signal $x$ is defined as
$$r = \min \{\Vert \hat{x}+x\Vert_2, \Vert \hat{x}-x\Vert_2 \}/\Vert x \Vert_2 .$$
We say it is a successful recovery when the relative estimation error $r$ satisfy that
$r\le 1e-2$ or the support is exactly recovered. The tests repeat independently for $100$ times to compute a successful rate. ``Aver Iter'' in the table \ref{tab:sparsity} and \ref{tab:dimension} means the average number of iterations for 100 times of tests. All the computations were performed on an eight-core laptop with core i7 6700HQ@3.50 GHz and 8 GB RAM using \texttt{MATLAB} 2018a.

\begin{algorithm}[hbt!]
   \caption{StormSpar}\label{alg:ithc}
   \begin{algorithmic}[1]
     \STATE Input: Normalized $A\in \mathbb{R}^{m\times n}$, $y$, sparsity level $s$, $\gamma \in (0,1)$,  small constant $\delta$, a random initial value $x^0$.
     \FOR {$\ell=1,2,...$}
     \STATE Randomly selected $\lfloor \gamma m\rfloor$ rows of $A$ and $y$,  denote the index as $i^{\ell}$, to form  $A^{\ell} = A(i^{\ell},:)$  $y^{\ell} = y(i^{\ell})$.
     \STATE Compute $p^{\ell} = \mathrm{sign}(A^{\ell}x^{\ell-1}), \tilde{y}^{\ell} = p^{\ell}\odot y^{\ell}$.
     \STATE Get $x^{\ell}$ by solving $\min_{x, \|x\|_0\leq  s} \frac{1}{2}\|A^{\ell} x - \tilde{y}^{\ell}\|^2$
     via Algorithm \ref{alg:htp} (HTP).
      \STATE Check stop criteria ${\lVert x^{\ell }-x^{\ell -1}\rVert}\le \delta$.
     \ENDFOR
     \STATE Get the first $s$ position of  $x^{\ell}$ and refit on it as output.
   \end{algorithmic}
\end{algorithm}

\begin{algorithm}[hbt!]
   \caption{HTP solving \eqref{pro:cs}}\label{alg:htp}
   \begin{algorithmic}[1]
     \STATE Input: Initialization: $k=0, x^{0}=0; $
     \FOR {$k=1,2,...$}
     \STATE $S^{k}\leftarrow$ \{indices of s largest entries of $x^{k-1}+\mu (A^\ell)^t( \tilde{y}^\ell-A^\ell x^{k-1})$\};\\
     \STATE Solve $x^{k}\leftarrow \mathrm{argmin}_{\mathrm{supp}(x)\subset S^{k}}\lVert A^\ell x-\tilde{y}^\ell\lVert_{2}.$
     \ENDFOR
    \end{algorithmic}
\end{algorithm}

\subsection{Examples}
\begin{figure}
  \centering
 \includegraphics[trim =0cm 0cm 0cm 0cm, clip=true,width=0.3\textwidth]{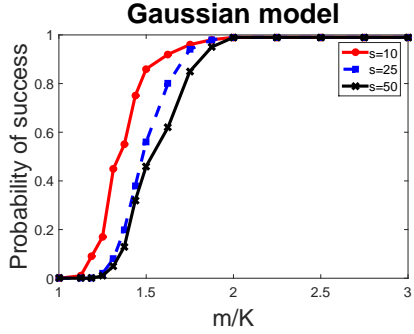}
 \caption{\label{sample} The probability of success in recovery v.s. sample size $m/K$ for Gaussian model, $K=\lfloor(s(\mathrm{log} \ n+\mathrm{log}\frac{1}{0.01}))\rfloor$ which is $115,287,575$ with respect to sparsity $s=10,25,50$, signal dimension $n=1000$, noise level $\sigma=0.01$.}
\end{figure}

\begin{figure}
  \centering
 \includegraphics[trim =0cm 0cm 0cm 0cm, clip=true,width=0.3\textwidth]{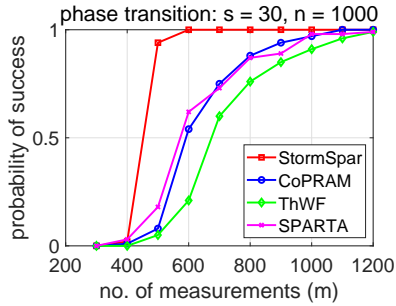}
 \caption{\label{samplecom} The probability of success in recovery for different algorithms in terms of changing sample size, dimension $n=1000$, sparsity $s=30$ and the model is noise free.}
\end{figure}

\textbf{Example 1} First we examine the effect of sample size $m$ to the probability of successful recovery in Algorithm 1. The dimension of the signal $x$ is $n=1000$. \\
a.) When we set sparsity to be $s=10,25,50$, Fig. \ref{sample} shows how the successful rate changes in terms of the sample size $m$.
In this experiment, we fix a number $K=\lfloor(s(\mathrm{log}\, n+\mathrm{log}\, \frac{1}{0.01}))\rfloor$, which is $115,287,575$ with respect to the sparsity $10,25,50$. Then we compute the probability of success when $m/K$ changes: for each $s$ and each $m/K=1,1.25,\cdots,3$, we run our algorithm for $100$ times. We find it that when the sample size is in order $O(s\,\mathrm{log}\,n )$ in this setting, we can recover the signal with high possibility.\\

b.) We compare StormSpar  to some existing algorithm, i.e. CoPRAM\cite{jagatap2019sample}, Thresholded Wirtinger Flow(ThWF)\cite{cai2016optimal} and SPArse truncated Amplitude flow (SPARTA)\cite{wang2016sparse}. The sparsity is set to be $30$ and the model is noise free.  Fig. \ref{samplecom} shows the successful rate comparison in terms of sample size, the results are obtained by averaging the results of $100$ trials. We find it that StormSpar  requires more iterations and more cpu time than these algorithms which requires initialization. But StormSpar  achieves better accuracy with less sample complexity.

\begin{figure}[htb!]
  \centering
 \includegraphics[trim =0cm 0cm 0cm 0cm, clip=true,width=0.3\textwidth]{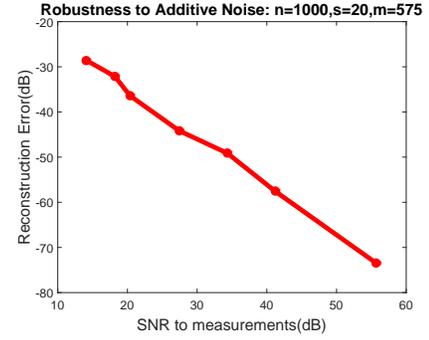}
 \caption{\label{noise} The reconstruction error v.s. SNR to measurements for Gaussian model, $m=\lfloor(2.5s(\mathrm{log} \ n+\mathrm{log}\frac{1}{0.01}))\rfloor=575$ with sparsity $s=20$, signal dimension $n=1000$ and several noise level, i.e. SNR to measurements.}
\end{figure}

\textbf{Example 2} Fig. \ref{noise} shows that StormSpar  is robust to noise. We set $n=1000,s=20$, and $m=\lfloor(2.5s(\mathrm{log}\ n+\mathrm{log}\frac{1}{0.01}))\rfloor (=575)$. The noise we added is i.i.d. Gaussian, and the noise level is shown by signal-to-noise ratios (SNR), we plot the corresponding relative error of reconstruction in the Fig. \ref{noise}. The results are obtained by average of $100$ times trial run.

\begin{figure}
  \centering
  \includegraphics[trim =0cm 0cm 0cm 0cm, clip=true,width=0.3\textwidth]{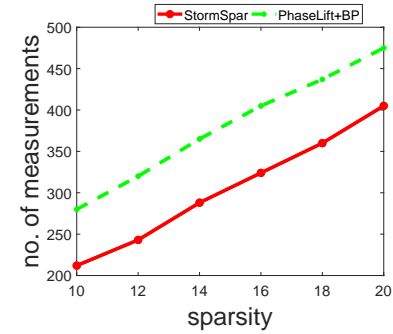}
\caption{\label{sample_sparsity1} Comparison of Minimum number of measurements required for Gaussian model, signal dimension $n=1000$ and free of noise.}
\end{figure}
\begin{figure}[htb!]
  \centering
  \includegraphics[trim =0cm 0cm 0cm 0cm, clip=true,width=0.3\textwidth]{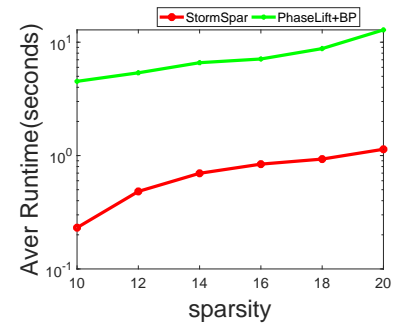}
 \caption{\label{sample_sparsity2} Comparison of efficiency for Gaussian model, signal dimension $n=1000$ and free of noise.}
\end{figure}

\textbf{Example 3} We compare StormSpar  with a two-stage method Phaselift+BP proposed in \cite{iwen2017robust}, which has been shown to be more efficient than the standard SDP of \cite{ohlsson2012cprl}. The dimension of data is set to be $n=1000$. The comparison are two-folder. Firstly, we compare the minimum number of measurements required to be sample size which gives successful recovery rate higher than $95\%$ for different sparsity level, the result can be found in Figure \ref{sample_sparsity1}. Secondly the average computational time is given in Figure \ref{sample_sparsity2}, where $m=\lfloor (2.5s(\mathrm{log}\, n+\mathrm{log}\, \frac{1}{0.01})) \rfloor$.



\textbf{Example 4} Let $m=O(s \mathrm{log} \, n)$, we test for different sparse levels and differen dimensions.
In Table \ref{tab:sparsity}, we fix dimension $n=2000$, and the sample size is chosen to be $m=\lfloor (2.5s(\mathrm{log}\, n+\mathrm{log}\, \frac{1}{0.01})) \rfloor$. The sparsity level chages from $5$ to $100$, we find the algorithm can successfully recover the sparse signal in most case, and the iteration number is very stable.

In Table \ref{tab:dimension}, the sparsity level is fixed by $s =10$, the sample size is $m=\lfloor (2.5s(\mathrm{log}\, n+\mathrm{log}\, \frac{1}{0.01})) \rfloor$ for dimension $n$ from $100$ to $10000$. We find the algorithm can successfully recover the sparse signal in most cases, and the number of iteration dependent on the dimension $n$ in a sublinear manner.

%
%

\begin{table}[htb!]
\centering
\caption{Numerical results for sparsity test, with random sampling
$A$ of size $ n\times m$, $n=2000$, $m=\lfloor (2.5s(\mathrm{log}\, n+\mathrm{log}\, \frac{1}{0.01})) \rfloor$,
$s$ is the sparsity, with $\sigma=\mbox{0.01}$, and Aver Iter$=\lfloor$ average number of iterations for 100 times of test$\rfloor$ . }\label{tab:sparsity}
\begin{tabular}{cccccp{0.3cm}p{0.3cm}c}
\hline\hline
\multicolumn{1}{c}{Dimension $n$} & \multicolumn{1}{c}{Sparsity $s$} & \multicolumn{1}{c}{Sample $m$} & \multicolumn{1}{c}{Successful Rate}& \multicolumn{1}{c}{Aver Iter}  \\
 \hline
    $2000$          &5              &  152   &98\%    &109                  \\
    $2000$          &10              & 305   &99\%   &229                  \\
    $2000$          &15              &  457   &99\%   &359                   \\
    $2000$          &20              &  610   &98\%   &395                   \\
    $2000$          &25              &  762   &100\%  &407                   \\
    $2000$          &30              &  915   &99\%   &403                  \\
    $2000$          &35              & 1068   &100\%   &482                   \\
    $2000$          &40              & 1220   &100\%   &331                   \\
    $2000$          &45              & 1373   &100\%   &305                   \\
    $2000$          &50              & 1525   &100\%   &324                   \\
    $2000$          &75              & 2288   &100\%   &289                   \\
    $2000$          &100              &3051   &100\%  &285                  \\
 \hline\hline
\end{tabular}
\end{table}

\begin{table}[htb!]
\centering
\caption{Numerical results for different dimensions, with random sampling
$A$ of size $n\times m$, $m=\lfloor (2.5s(\mathrm{log}\, n+\mathrm{log}\, \frac{1}{0.01})) \rfloor$,
$s$ is the sparsity, with $\sigma=\mbox{0.01}$, and Aver Iter$=\lfloor$ average number of iterations for 100 times of test$\rfloor$ . }\label{tab:dimension}
\begin{tabular}{cccccp{0.3cm}p{0.3cm}c}
\hline\hline
\multicolumn{1}{c}{Dimension $n$} & \multicolumn{1}{c}{Sparsity $s$} & \multicolumn{1}{c}{Sample $m$} & \multicolumn{1}{c}{Successful Rate}& \multicolumn{1}{c}{Aver Iter}  \\
 \hline
    $100$          &10              &  230   &98\%   &38                  \\
    $200$          &10              &  249   &99\%   &46                  \\
    $300$          &10              &  257   &100\%  &56                   \\
    $400$          &10              &  264   &100\%  &72                   \\
    $500$          &10              &  270   &100\%  &93                   \\
    $750$          &10              &  280   &100\%  &123                  \\
    $1000$          &10              & 287   &100\%  &157                   \\
    $1500$          &10              & 297   &99\%   &192                   \\
    $2000$          &10              & 305   &99\%   &229                   \\
    $3000$          &10              & 315   &99\%   &298                   \\
    $4000$          &10              & 322   &98\%   &508                  \\
    $5000$          &10              & 328   &97\%   &748                  \\
    $7500$          &10              & 338   &95\%   &1142                  \\
    $10000$          &10             & 345   &96\%   &1271                  \\

\hline\hline
\end{tabular}
\end{table}


\section{Conclusion}

In this paper, we have proposed a novel algorithm (StormSpar) for the sparse phase retrieval. StormSpar  start with a random initialization and employ a alternating minimizing method for a changing objective function. The subproblem $\min_{x, \|x\|_0\leq  s} \frac{1}{2}\|A^{\ell} x - \tilde{y}^{\ell}\|^2$ is a standard compressed sensing problem, which can be solved by HTP method. Numerical exampls show that the proposed algorithm requires only $O(s\, \mathrm{log}\, n)$ samples to recover the $s$-sparse signal with a random initial guess.
\section*{Acknowledgements}
The research of J.-F.  Cai is partially supported by Hong Kong Research Grant Council (HKRGC) grant GRF 16306317.
The research of Y. Jiao is partially supported by
National Science Foundation of  China (NSFC) No. 11871474 and 61701547.
The research of X. Lu is partially supported by NSFC Nos.  91630313 and  11871385.

\bibliographystyle{IEEEtran}
\bibliography{phaseretrieval}
\end{document}